\documentclass[a4paper]{gretsi}   	
\usepackage{subcaption}
\usepackage{graphicx}									
\usepackage{amssymb}
\usepackage{amsmath}
\usepackage{xcolor}
\usepackage[english,french]{babel} 
\usepackage[T1]{fontenc}
\usepackage[utf8]{inputenc}

\DeclareMathOperator*{\argmin}{arg\,min}

\usepackage{algorithm}
\usepackage{algorithmic}
\usepackage{slashbox} 
\usepackage{multirow}
\usepackage{url}
\usepackage[colorinlistoftodos]{todonotes}

\DeclareMathSymbol{\R}{\mathalpha}{AMSb}{"52}
\newcommand{\x}{\mathbf{x}}
\newcommand{\y}{\mathbf{y}}
\newcommand{\db}{\mathbf{d}}

\newcommand{\ub}{\mathbf{u}}

\newcommand{\z}{\mathbf{z}}

\newcommand{\prox}{\mbox{prox}}

\usepackage{tabulary}
\newcolumntype{C}[1]{>{\centering\arraybackslash}p{#1}}   
\newcolumntype{L}[1]{>{\raggedright\arraybackslash}p{#1}}
\usepackage{colortbl}  

\definecolor{aliceblue}{rgb}{0.94, 0.97, 1.0}
\definecolor{beaublue}{rgb}{0.74, 0.83, 0.9}

\title{Une véritable approche $\ell_0$ pour l'apprentissage de dictionnaire} 
\usepackage{times}	

\author{\coord{Yuan}{Liu}{},
        \coord{Stéphane}{Canu}{},
    \coord{Paul}{Honeine}{},
    \coord{Su}{Ruan}{}}

\address{\affil{}{Normandie Univ, INSA Rouen, UNIROUEN, UNIHAVRE,  LITIS; \\
         Avenue de l'Université, 76801 Saint-Étienne-du-Rouvray Cedex, France}}
         
\email{yuan.liu@insa-rouen.fr,
stephane.canu@insa-rouen.fr\\
paul.honeine@univ-rouen.fr, su.ruan@univ-rouen.fr}

\frenchabstract{Ces derniers temps, les modèles parcimonieux ont suscité un vif intérêt  
de part leur capacité à sélectionner automatiquement un modèle simple parmi une grande collection. 
Ils se sont notamment révélés être utiles pour l'apprentissage de dictionnaire. 
La manière classique d'exprimer la parcimonie dans ce cadre, consiste à utiliser la norme $\ell_0$ qui permet de compter, et donc de contrôler, le nombre de composantes d'un modèle.
Malheureusement, les problèmes d'optimisation associés s'avèrent être non convexes et NP-difficiles,  
ce qui a justifié la recherche de relaxations pour obtenir une bonne approximation de la solution globale du problème.
A l'inverse, nous montrons dans cet article que,  en dépit de sa complexité, 
ce problème l'apprentissage de dictionnaire avec la norme $\ell_0$ 
peut aujourd'hui être traité efficacement. 
L'idée est de reformuler le problème comme un programme quadratique mixte en nombres entiers (MIQP) et d'utiliser un logiciel d'optimisation pour obtenir l'optimum global du problème. 
La principale difficulté de cette approche étant le temps de calcul, nous proposons deux méthodes pour le  réduire. 
L'application de notre méthode d'apprentissage de dictionnaire MIQP à un problème de débruitage d'images démontre sa faisabilité.}

\englishabstract{Sparse representation learning has recently gained a great success in signal and image processing, thanks to recent advances in dictionary learning. To this end, the $\ell_0$-norm is often used to control the sparsity level. Nevertheless, optimization problems based on the $\ell_0$-norm are non-convex and  NP-hard. For these reasons, relaxation techniques have been attracting much attention of researchers, by priorly targeting approximation solutions ({\em e.g.} $\ell_1$-norm, pursuit strategies). On the contrary, this paper considers the exact $\ell_0$-norm optimization problem and proves that it can be solved effectively, despite of its complexity. The proposed method reformulates the problem as a Mixed-Integer Quadratic Program (MIQP) and gets the global optimal solution by applying existing optimization software. Because the main difficulty of this approach is its computational time, two techniques are introduced that improve the computational speed. Finally, our method is applied to image denoising which shows its feasibility and relevance compared to the state-of-the-art.}

\begin{document}
\maketitle
\section{Introduction}

L'apprentissage de représentations parcimonieuses a été largement considéré 
avec succès dans les domaines du traitement du signal, des images et en vision, notamment pour des applications de débruitage d'image \cite{elad2006image,bao2014l0}, d'inpainting \cite{mairal2009online} et de classification \cite{zhang2010discriminative}, 
pour n'en citer que quelques-unes.
%
Il consiste à modéliser les données par une combinaison linéaire de quelques éléments d'un dictionnaire.
Au delà d'un dictionnaire prédéfini, 
nous considérons ici l'apprentissage du dictionnaire pour une représentation adaptée aux données disponibles.

L'apprentissage nécessite alors l'estimation jointe des éléments du dictionnaire et de leur coefficient de pondération. En opérant une procédure d'optimisation alternée, les éléments du dictionnaire peuvent être estimés facilement à chaque itération par moindres carrés ou descente de gradient stochastique \cite{mairal2009online}. 
L'estimation des coefficient de pondération, dite codage parcimonieux ({\em sparse coding} en anglais), est un problème non convexe et NP-difficile à cause de la contrainte de type $\ell_{0}$ pour imposer la parcimonie. 
Afin de surmonter cette difficulté, deux principales approches ont été mises en {\oe}uvre pour relaxer la contrainte $\ell_{0}$. La première considère une solution approchée par 
poursuite séquentielle \cite{elad2006image,15.sparse.eigen}. 
La seconde approche opère en remplaçant la norme $\ell_{0}$ par sa relaxation convexe~: la norme~$\ell_{1}$.
Les méthodes les plus connues sont la méthode bayésienne \cite{olshausen1997sparse}, 
la méthode \textit{K-SVD} \cite{aharon2006rm}  
et la méthode proximale \cite{bao2014l0}.

Le présent article vise à résoudre d'une manière exacte le problème de l'apprentissage de dictionnaire, c'est à dire en considérant la norme $\ell_{0}$ sans aucune relaxation. Pour ce faire, nous reformulons le problème d'optimisation pour le résoudre par programme quadratique mixte en nombres entiers (MIQP). Ainsi, les récentes avancées théoriques en optimisation sont-elles exploitées avec les améliorations d'implémentation qui les accompagnent. Nous proposons aussi deux techniques pour réduire le coût calculatoire, d'une part avec la mise en place de nouvelles contraintes pour renforcer la formulation et d'autre part en initialisant par une méthode proximale. 
Nous montrons que ces différentes contributions permettent la résolution exacte de l'apprentissage parcimonieux pour le débruitage d'images. Les résultats obtenues corroborent une récente étude sur la tolérance élevée de MIQP à la présence du bruit \cite{bourguignon2016exact}.

L'article est organisé comme suit. 
Le problème de représentation parcimonieuse est présenté et la méthode d'optimisation exacte est décrite dans la section 2.  
La section 3 montre la pertinence de la méthode proposée en débruitage d'image, et la dernière section conclut l'article.

%

\section{Représentation parcimonieuse}
\subsection{Énoncé du problème}

Soit $Y= [\y_{1}, \dots ,  \y_{\ell} ] \in \R^{n \times \ell}$, une matrice  contenant $\ell$ signaux $\y_{i}$, $i=1,\dots,\ell$, de dimension $n$. On suppose que $Y = M+B$ où $M$ et $B$ sont deux matrices modélisant respectivement les parts d'information et de bruit inconnu, contenues des signaux.
La représentation parcimonieuse de $Y$, consiste à trouver une matrice $X = [\x_{1}, \dots , \x_{\ell}] \in \R^{ p\times \ell}$ parcimonieuse ({\it i.e.}, avec seulement quelques termes non nuls) et un dictionnaire $D = [\db_{1}, \dots,  \db_{p}] \in \R^{n \times p}$ tels que  $M = DX$. 
Les éléments $\db_{i}$, $i=1, \dots , p$, sont appelés atomes et $D$ appartient à l'espace $\mathcal{D} = \{D \in \R^{n \times p},  \db_{j}^{T}\db_{j} \leq 1, j = 1, \dots, p\}$. 
L'estimation jointe de $X$ et de $D$ peut s'écrire comme un problème de minimisation du risque empirique régularisé~: 
%
\begin{equation}
\label{eq:min_risk_regul}
\displaystyle \min_{\substack{D\in \mathcal{D}\\ X \in \R^{p \times \ell}\!\!\!\!\!}} ~\; \frac{1}{\ell} \sum_{i=1}^{\ell} \bigl( \tfrac{1}{2} \Vert \y_{i} - D \x_{i} \Vert_{2}^2 +\lambda \Omega(\x_{i})\bigr).
\end{equation}
Le premier terme représente l'erreur de reconstruction et le second  la régularisation, qui comprend un opérateur de régularisation $\Omega(\x_{i})$. Le paramètre $\lambda > 0$ contrôle le compromis entre la fidélité aux données et  la parcimonie. 
Une manière classique d'aborder ce problème \eqref{eq:min_risk_regul} d'estimation jointe de $D$ et $X$ consiste à utiliser une procédure de relaxation alternée en deux phases \cite{aharon2006rm}. 
La première phase, dite de codage parcimonieux  (\textit{sparse coding}), consiste à estimer $X$ en supposant $D$ connu.
 Les méthodes les plus utilisées sont celle de Gauss Seidel   \cite{tropp2007signal} et de descente de gradient \cite{mairal2014sparse}. 
La seconde phase, dite d'apprentissage de dictionnaire, consiste à estimer $D$ en supposant $X$ connu. Les algorithmes les plus utilisés sont ceux des moindres carrés et de descente de gradient stochastique \cite{mairal2009online}.

Dans ce travail nous nous intéressons au cas où la régularisation est de type $\ell_{0}$,
($\Omega(\x) = \Vert \x \Vert_{0}$), ce qui permet de contrôler explicitement le nombre de termes non nuls du vecteur $\x$. 
En pratique, deux  formulations de minimisation sous contraintes analogues à \eqref{eq:min_risk_regul}
 peuvent être utilisées. La première s'écrit, pour un $T>0$ majorant le nombre de composantes non nulles~: 
%
%
\begin{equation}
\label{eq:L0problem}
\displaystyle \min_{\substack{D \in \mathcal{D}\\ \x_{i} \in {\R}^p}} \tfrac{1}{2} \Vert \y_{i}-D\x_{i} \Vert_{2}^2  
~\text{ avec }~
\Vert \x_{i} \Vert_{0} \leqslant T,\; i=1,\ldots,n.
\end{equation}
%
%
La seconde s'écrit, pour $\varepsilon > 0$ représentant le niveau de bruit~: 
\begin{equation}
\label{eq3}
\displaystyle \min_{\substack{D \in \mathcal{D}\\ \x_{i} \in \R^{p}}} \Vert \x_{i} \Vert_{0}  
~\text{ avec }~ 
\tfrac{1}{2} \Vert \y_{i}-D\x_{i} \Vert_{2}^2 \leqslant \varepsilon,\; i=1,\ldots,n.
\end{equation}
%
%
%
Quelle que soit la formulation choisie, \eqref{eq:min_risk_regul}, \eqref{eq:L0problem} ou \eqref{eq3},
dans le cadre de l'approche de minimisation alternée que nous nous proposons d'utiliser, à cause de la la norme $\ell_0$, le problème d'optimisation lié à la phase de codage parcimonieux est non-convexe et NP-difficile. 
Constatant cette difficulté, la plupart des travaux dans le domaine proposent 
de relaxer le problème en remplaçant la norme $\ell_0$ par un terme convexe comme la norme $\ell_1$ \cite{mairal2009online} ou d'utiliser un algorithme approché comme celui de poursuite \cite{elad2006image}.
Dans cet article, nous proposons un nouvel algorithme basé sur la programmation quadratique mixte binaire,  permettant 
de résoudre de manière exacte le problème   de codage parcimonieux avec la norme $\ell_0$ associé à l'équation~\eqref{eq:L0problem}.


\subsection{Optimisation globale du problème}

%
\subsubsection{Programmation quadratique mixte (MIQP)}
Une manière de traiter la norme $\ell_0$ qui apparaît dans la phase de {\it sparse coding} associée au problème d'optimisation~\eqref{eq:L0problem}, 
consiste à réécrire le problème sous une forme standard 
que l'on sait résoudre avec les logiciels 
d'optimisation d'aujourd'hui.

Cette réécriture peut s'effectuer grâce à l'introduction de variables binaires.
L'idée est d'associer à tous les éléments $x_i$ du vecteur  $\x$, une variable binaire $z_i$ 
égale à 0 si $x_i=0$ et égale à un sinon. Cela revient à imposer la relation logique suivante, composante par composante
 %
 \begin{equation} \label{eq:logic}
 z_{i} = 0  \;  \Longleftrightarrow \;  x_{i} = 0, \qquad  i = 1, \dots, p.
\end{equation}
A l'aide de cette nouvelle variable binaire, la contrainte de parcimonie $\Vert \x \Vert_{0} \leqslant T$ 
peut s'exprimer sous la forme
%
\begin{equation}
\sum_{i=1}^{p} z_{i} \leqslant T,
\end{equation}
de sorte que le problème de {\it sparse coding} associé à \eqref{eq:L0problem} se réécrive~: 
%
\begin{equation}
\begin{array}{cl}
\displaystyle \min_{\substack{\x \in \R^p \\ \z\in \{0,1\}^p}} & \tfrac{1}{2}  \Vert \y-D\x \Vert_{2}^2\\
\text{ avec } & z_{i} = 0  \;  \Leftrightarrow \;  x_{i} = 0, \quad  i = 1, \dots, p\\
 &\bold{1}_{p}^{T}\z \leqslant T, 
\end{array}
\label{eq:reformulation2}
\end{equation}
où $\bold{1}_{p}$ est un vecteur de 1 de dimension $p$. 

Si la relation logique \eqref{eq:logic} est traitée par certains logiciels, il est parfois préférable de l'éliminer explicitement. Ce peut être réalisé par l'intermédiaire de l'introduction d'une contrainte de type {\it big~$M$}. 
Supposons que l'on connaisse un réel $M > 0$ suffisamment grand, de sorte que, si $\x^\star$ est solution du problème  \eqref{eq:L0problem}, alors   $\Vert \x^\star \Vert_{\infty} < M$. 
Dans ce cas, imposer les contraintes  \eqref{eq:logic} revient à poser
%
\begin{equation}
-z_{i}M \leqslant  x_{i}  \leqslant z_{i}M, \quad  i \in 1, \dots, p.
\label{ineq}
\end{equation}
%
La formulation {\it big $M$} du problème \eqref{eq:L0problem} est donc, pour $M >0$ et $T>0$ donnés~:
\begin{equation}
\begin{array}{cl}
\displaystyle \min_{\substack{\x \in \R^p \\ \z\in \{0,1\}^p}} & \tfrac{1}{2}  \Vert \y-D\x \Vert_{2}^2\\
\text{avec} & -\z M \leqslant  \x  \leqslant \z M\\
 & \; \; \bold{1}_{p}^{T}\z \leqslant T.
\end{array}
\label{eq:reformulation}
\end{equation}
%
%
Selon \cite{bourguignon2016exact}, la contrainte dans \eqref{eq:reformulation} est équivalent à celle de \eqref{eq:L0problem}. De plus, $D$ doit satisfaire la propriété d'\textit{Unique Representation Property} \cite{gorodnitsky1997sparse} qui assure l'unicité de la solution.
Analysons maintenant ce problème d’optimisation. D'abord, sa fonction objectif est quadratique. 
Ensuite, il comporte deux types des variables $\x$ et $\z$ respectivement continues et entières~: c'est ce qu'on appelle un problème d'optimisation mixte en nombre binaires (ou plus généralement en nombre entiers). 
Enfin, les contraintes, quand à elles, sont linéaires.
Ce type de problème est connu sous le nom de programme quadratique mixte en nombre binaires ou en anglais  {\it mixed binary (integer) quadratic programming} (MIQP). 
%
Ce problème MIQP \eqref{eq:reformulation}
 peut être résolu exactement sur les images qui nous intéressent en utilisant un logiciel d'optimisation comme CPLEX ou GUROBI. 
 %

\subsubsection{L'introduction de contraintes complémentaires}

En programmation mixte, il est bien connu qu'une <<~bonne~>> formulation des contraintes peut grandement accélérer les performances d'un solveur \cite{neumaier2004safe}.
%
%
Notamment, si dans le meilleur des cas on arrive à exprimer des contraintes définissant l'enveloppe convexe du domaine admissible, 
alors la solution optimale d'un programme mixte est la même que celle de sa relaxation continue \cite{hoffman2013integer}. Malheureusement, l'obtention de cette enveloppe convexe est un problème NP-difficile. En revanche, l'enveloppe convexe des contraintes sur les 
variables continues
$$
\mathcal{C} = \Bigl\{\x \in \R^p\; \bigl|\bigr.\;    
\z \in \{0,1\}^{p},\;
\sum_{j=1}^{p}  z_j \leq T,\; |\x_j | \leq  z_j T, \Bigr\},
$$
peut s'exprimer à l'aide des normes un et infinies de $\x$ comme~:
$$
\Bigl\{\x \in \R^p\; \bigl|\bigr.\;    
\|\x\|_1 \leqslant T M ,\; ||\x \|_\infty  \leqslant   M \Bigr\}.
$$
Nous proposons  d'ajouter ces contraintes au problème~\eqref{eq:reformulation} pour obtenir le problème suivant équivalent mais mieux structuré~:
\begin{equation}
\begin{array}{cl}
\displaystyle \min_{\substack{\x \in \R^p \\ \z\in \{0,1\}^p}} & \tfrac{1}{2}  \Vert \y-D\x \Vert_{2}^2\\
\text{avec} & -\z M \leqslant  \x  \leqslant \z M\\
 & \; \; \bold{1}_{p}^{T}\z \leqslant T \\
 & \; \|\x\|_1 \leqslant T M\\
 & ||\x \|_\infty  \leqslant   M.
\end{array}
\label{eq:add_constrints}
\end{equation}
Cette formulation permet au solveur d'obtenir la même solution que \eqref{eq:reformulation}, mais plus rapidement.

Il reste que les performances des solveurs sur cette formulation sont très sensibles au choix de la constante  $M$. 
Nous allons maintenant voir comment régler ce paramètre 
en utilisant une procédure proximale du premier ordre permettant en plus, d'obtenir une bonne initialisation des variables à optimiser.

\subsubsection{Initialisation par la méthode du gradient proximal}

L'algorithme du gradient proximal est une méthode du premier ordre permettant d'obtenir rapidement une solution locale du problème \eqref{eq:L0problem}. Cette solution peut être utilisée comme une bonne initialisation des variables à optimiser et du paramètre $M$, permettant aux solveurs d'accéder plus rapidement au minimum globale du problème \cite{atamturk2005integer}.
L’approche proximale consiste à minimiser itérativement une succession de majorations de la fonction objectif.
Elle est construite à partir de  
l'opérateur proximal associé à la contrainte $\|\x\|_0 \leqslant T$~:
$$
\begin{array}{rl}
 \prox_T : \R^p &\longrightarrow \R{^p} \\
\x &\longmapsto \prox_T(\x) =  \displaystyle \argmin_{\|\ub\|_0 \leqslant T} \; \tfrac{1}{2} \|\ub-\x\|^2.
\end{array}
$$
Il est facile de voir que la solution de ce problème 
est donnée par les $T$ plus grandes valeurs absolues des composantes du vecteur $\x$, soit
$$
 \prox_T(\x) = 
 \left\{\begin{array}{cl}
 x_j & \mbox{ si }  j \in \{(1),\dots,(T)\}  \\
 0 & \displaystyle   \mbox{ sinon,} %
 \end{array}
 \right.
$$ 
où $(j)$ est la suite d'indices tels que  $|x_{(1)}| \geqslant \dots \geqslant |x_{(p)}| $.  
L'algorithme de descente de gradient proximal
consiste alors à mettre en \oe uvre, pour un pas $\rho$, les itérations suivantes \cite{bao2014l0}~:
%
$$
\displaystyle \x^{k+1} \in \prox_{T}\bigl(\x^{k}- \rho D^T(D\x^{k} - \y)\bigr).
$$
Lorsque le pas  $\rho$ est bien choisi, il est possible de démontrer que l'algorithme proximal converge vers un minimum local. 
Soulignons que, comme pour la plupart des méthodes itératives, il existe de nombreuses variantes de l'algorithme permettant  d'accélérer la convergence. 

Connaissant $\x^\star$ le point de convergence de l’algorithme proximal, il est possible d'en déduire une initialisation pour le vecteur $\z$ et le paramètre $M$. 
Par exemple, pour un $\varepsilon > 0$ donné, on peut initialiser $\z$ avec $z_j = 0$ si $|x_j^\star| \leqslant \varepsilon$ et $z_j = 1$ sinon.
La constante $M$ peut être choisie telle que $M = (1+\alpha) \Vert \x^{\star} \Vert_{\infty}$ avec   $\alpha> 0$ choisi le plus petit possible. 

Par la résolution exacte de \eqref{eq:L0problem} pour déterminer le codage parcimonieux, la convergence de notre algorithme peut être assuré selon l'analyse conduite dans \cite{aharon2006rm}.

%


\section{Résultats expérimentaux}
 

Le but de nos expériences est de comparer les performances de notre méthode MIQP avec celles des algorithme de référence sur une tâche de débruitage. 
Nous avons travaillé sur cinq images naturelles de bonne qualité fréquemment utilisées et extraites  de \textit{Miscellaneous volumes of the USC-SIPI Image Database}\footnote{\scriptsize\url{http://sipi.usc.edu/database/database.php?volume=misc}} (Barbara, Cameraman, Elaine, Lena et Men). 

Nous avons construit la matrice $Y$ des données d'apprentissage à l'aide des cinq images simultanément
en utilisant, comme dans la littérature \cite{elad2006image}, des imagettes de taille $8 \times 8$ se chevauchant. Notre matrice $Y$ a donc pour dimension $n = 64$ et  $\ell >  3,\!5 \times 10^{4}$. 
Nous avons fixé expérimentalement le nombre d'atomes du dictionnaire à $p = 100$ et le coefficient de parcimonie à $T = 20$. 
Nous avons testé différents niveaux de bruit additif gaussien sur les images
en utilisant trois différents niveaux de bruit avec des valeurs d'écart type $ \sigma = 10, 20$ et $50$. 
Pour chaque expérience, nous  avons utilisé la procédure de relaxation alternée et nous avons itéré 30 fois les deux phases successives de codage parcimonieux et d'apprentissage de dictionnaire.

Pour la phase de codage parcimonieux, nous avons comparé notre approche MIQP avec les deux méthodes références de la littérature, 
 K-SVD \cite{elad2006image} et la méthode proximale \cite{bao2014l0}, toutes choses égales par ailleurs. Nous avons aussi comparé deux méthodes de reconstruction : la méthode directe où l'image est reconstruite par $DX$ et l'approche proposée par Elad \emph{et al.} \cite{elad2006image} où elle est estimée par une combinaison linéaire entre $Y$ et $DX$.

Nous avons réalisé nos expériences en Matlab sur un PC Dell T5500 à 8 c\oe urs. 
Le nombre d’itérations maximal de la méthode proximale a été fixé à 200. 
Nous avons utilisé GUROBI 7.0 pour résoudre les MIQP avec un temps d'exécution maximal de 50 secondes, un nombre d'itération maximal de 200. Nous avons aussi fixé $\alpha = 1,\!5$ pour des raisons de stabilité.


La \tablename~\ref{Table:PSNRresult} résume nos résultats. 
La première remarque est que, pour un fort niveau de bruit ($\sigma = 50$) et sur toutes les images testées, notre méthode MIQP donne de meilleurs résultats (le rapport signal sur bruit est plus grand) que les autres approches de codage parcimonieux (K-SVD et l’algorithme proximal seul). L'amélioration peut être quantifiée en moyenne par une augmentation de $1,\!79$ par rapport à la méthode proximale et de $3,\!73$ par rapport à K-SVD, soit un gain de près de 20 \%. 
Cela reste vrai quelle que soit la méthode de reconstruction utilisée. 
Nous avons aussi constaté que la méthode de reconstruction proposée par \cite{elad2006image} donne systématiquement de meilleurs résultats.
Cependant, pour un faible niveau de bruit ($\sigma = 10$), notre méthode ne réussit pas à améliorer les résultats de K-SVD alors que, pour un niveau intermédiaire ($\sigma = 20$) les résultats sont plus contrastés et dépendent de l'image considérée. Nous constatons que MIQP a tendance à moins bien se comporter sur des images de bonne qualité. 
D'une certaine manière, MIQP montre une meilleure capacité à éliminer le bruit via la parcimonie que les autres approches.

Quand nous comparons les résultats obtenus par chaque méthode pour différents niveaux de bruit, 
nous constatons aussi que ce sont ceux de MIQP qui diminuent le plus lentement que les autres méthodes lorsque $\sigma$ augmente. 
Soulignons enfin que, contrairement aux  algorithmes de référence de la littérature \cite{elad2006image,dang2016towards,bao2014l0} pour lesquels le niveau de bruit doit être connu, la méthode proposée débruite l'image sans des connaissance a priori autre que le niveau de parcimonie souhaité.
%


\section{Conclusion}

Dans cet article nous avons proposé, pour résoudre un problème de débruitage,  une véritable modélisation $\ell_0$ de la parcimonie et la reformulation du problème associé sous la forme d'un programme quadratique mixte en nombre entiers (MIQP). Nous avons montré que les logiciels d'optimisation disponibles aujourd'hui pouvaient donner la solution globale du problème en un temps raisonnable, ce qui nous a permis de traiter des problèmes de débruitage sur de vraies images par une méthode itérative d'apprentissage de dictionnaire,  exigeant à chaque itération la résolution de plus de 35 000 MIQP. 
Pour arriver à ce résultat, nous avons proposé deux techniques d'accélération du traitement des MIQP, la reformulation des contraintes pour  mieux structurer le problème, et l'initialisation efficace de la procédure grâce à un algorithme proximal.

Nos résultats démontent d'abord la faisabilité de notre approche. Les progrès conjugués des logiciels, du matériel et de la modélisation (notre compréhension de la nature du problème), permettent aujourd'hui d'utiliser la programmation mixte en nombre entiers pour résoudre des problèmes de traitement d'image. 
Cela ouvre la porte à une nouvelle approche des problèmes de modélisation de la parcimonie, puisqu'il est maintenant possible de la gérer explicitement grâce aux programmes mixtes dont on est en mesure de calculer la solution globale,  en dépit de leur caractère non convexe et NP difficile.

\begin{table}[tbp]
\caption{Résultats en terme de rapport signal sur bruit (PSNR) avec reconstruction standard de l'image (à gauche) et avec la reconstruction proposée dans Elad \emph{et al.} [6] (à droite). Les meilleurs résultats de chaque expérience sont en rouge.}
\begin{center}
\footnotesize
\renewcommand{\arraystretch}{1.3}
\begin{tabular}
{C{1.1cm}|C{.87cm}|C{.55cm}|C{.55cm}|>{\columncolor{beaublue}}C{.55cm}|C{.55cm}|C{.55cm}| >{\columncolor{beaublue}}C{.55cm}|}
\cline{2-8}
 & method & \multicolumn{3}{c|}{PSNR}& \multicolumn{3}{c|}{PSNR \cite{elad2006image}} \\
\hline
\multicolumn{1}{|C{1.1cm}|}{\backslashbox[0.8cm]{image\kern-2em}{$\sigma$}}&& 10 & 20 & 50 & 10 & 20 & 50 \\
\hline
\multicolumn{1}{|C{1.1cm}|}{\multirow{3}{*}{Barbara} }& K-SVD & \textcolor{red}{32,15} & \textcolor{red}{27,48} & 19,71&\textcolor{red}{33,6} & \textcolor{red}{28,25} & 20,03 \\
\cline{2-8}
\multicolumn{1}{|C{1.1cm}|}{}& proximal & 31,49 & 27,36 & 20,71 & 32,98 & 28,13 & 21,03 \\
\cline{2-8}
\multicolumn{1}{|C{1.1cm}|}{}& MIQP &26,42  &25,72  & \textcolor{red}{22,73} &27,91	&26,50 & \textcolor{red}{23,05} \\
\hline
\multicolumn{1}{|C{1.1cm}|}{\multirow{3}{*}{{\scriptsize Cameraman}}} & K-SVD &\textcolor{red}{ 29,53}	&26,45	&19,46	&\textcolor{red}{31,02}&	27,23	&19,78 \\
\cline{2-8}
\multicolumn{1}{|L{1.1cm}|}{}&proximal	&28,80	&\textcolor{red}{26,75}	&21,11	&30,30	&\textcolor{red}{27,53}	&21,43 \\
\cline{2-8}
\multicolumn{1}{|C{1.1cm}|}{}&MIQP	&25,90	&25,25	&\textcolor{red}{22,30}	&27,39	&26,03	&\textcolor{red}{22,62} \\
\hline
\multicolumn{1}{|C{1.1cm}|}{\multirow{3}{*}{Elaine}} & K-SVD	&32,93	&27,45	&19,73	&34,42	&28,52	&20,05 \\
\cline{2-8}
\multicolumn{1}{|c|}{}&proximal	&\textcolor{red}{33,14}	&28,99	&22,87	&\textcolor{red}{34,63}	&29,77	&23,19 \\
\cline{2-8}
\multicolumn{1}{|c|}{}&MIQP	&30,88	&\textcolor{red}{29,09}	&\textcolor{red}{24,20}	&32,38	&\textcolor{red}{29,87}	&\textcolor{red}{24,52} \\
\hline
\multicolumn{1}{|c|}{\multirow{3}{*}{Lena}} & K-SVD	&33,61	&27,91	&19,79	&35,10	&28,69	&20,11 \\
\cline{2-8}
\multicolumn{1}{|c|}{}&proximal	&\textcolor{red}{34,08}	&\textcolor{red}{29,52}	&22,12	&\textcolor{red}{35,57}	&\textcolor{red}{30,30}	&22,44 \\
\cline{2-8}
\multicolumn{1}{|c|}{}&MIQP	&30,82	&29,07	&\textcolor{red}{24,20}	&32,31	&29,85	&\textcolor{red}{24,52} \\
\hline
\multicolumn{1}{|c|}{\multirow{3}{*}{Men}} & K-SVD	&\textcolor{red}{31,95}	&27,36	&19,68	&\textcolor{red}{33,45}	&28,14	&20,00 \\
\cline{2-8}
\multicolumn{1}{|c|}{}&proximal	&31,62	&\textcolor{red}{28,20}	&21,26	&33,11	&\textcolor{red}{28,98}	&21,58 \\
\cline{2-8}
\multicolumn{1}{|c|}{}&MIQP	&28,47	&27,38	&\textcolor{red}{23,59}	&29,97	&28,16	&\textcolor{red}{23,91} \\
\hline
\end{tabular}
\end{center}
\label{Table:PSNRresult}
\end{table}

\bibliographystyle{abbrv}
{\scriptsize\bibliography{Biblio}}


\end{document}